# MKIoU Loss: Towards Accurate Oriented Object Detection in Aerial Images

Xinyi Yu, Jiangping Lu, Mi Lin, Linlin Ou*

Zhejiang University of Technology, Hangzhou, China

## Abstract

Oriented bounding box regression is crucial for oriented object detection. However, regression-based methods often suffer from boundary problems and the inconsistency between loss and evaluation metrics. In this paper, a modulated Kalman IoU loss of approximate SkewIoU is proposed, named MKIoU. To avoid boundary problems, we convert the oriented bounding box to Gaussian distribution, then use the Kalman filter to approximate the intersection area. However, there exists significant difference between the calculated and actual intersection areas. Thus, we propose a modulation factor to adjust the sensitivity of angle deviation and width-height offset to loss variation, making the loss more consistent with the evaluation metric. Furthermore, the Gaussian modeling method avoids the boundary problem but causes the angle confusion of square objects simultaneously. Thus, the Gaussian Angle Loss (GA Loss) is presented to solve this problem by adding a corrected loss for square targets. The proposed GA Loss can be easily extended to other Gaussian-based methods. Experiments on three publicly available aerial image datasets, DOTA, UCAS-AOD, and HRSC2016, show the effectiveness of the proposed method.

## 1. Introduction

As an emerging field, oriented object detection has recently seen rapid development. Different from general horizontal object detection, oriented object detection aims to locate arbitrary-oriented objects more accurately from complex backgrounds. It has been widely used in urban planning, surveying and mapping, and resource discovery. Although considerable progress has been made recently [1-5], challenges still exist, such as large aspect ratios, dense distribution, and arbitrary direction.

Most of the current detection algorithms are regression-based methods. Therefore, the design of the regression loss directly determines the performance of the detection. In generic horizontal detection, the common regression loss can be divided into two main types. One is Ln-norm-based, such as L1 loss, L2 loss, and Smooth L1 loss[6]. However, these methods optimize each parameter individually, ignore the correlation between the parameters, and are non-scale-invariant. Another is Intersection over Union (IoU)-based, such as GIoU[7], DIoU[8], CIoU[8], and EIoU[9]. The IoU-based losses are scale-invariant, and the IoU calculation process implicitly encodes the association between the parameters. Furthermore, as the primary evaluation metric in object detection, mean Average Precision (mAP) depends heavily on the IoU scores between the predicted box and ground truth box. Therefore, the IoU-based losses have a natural advantage over the Ln-norm losses. For horizontal object detection, the IoU can be easily calculated by the coordinates of the bounding box. However, for oriented object detection, due to the diversity of the two oriented boxes, the precise SkewIoU calculation process is too complex to be computed directly under existing deep learning frameworks[10]. Moreover, the boundary problems caused by the periodicity of the angle and the confusion of edges always affect the performance of the detection. Therefore, the design of loss function for oriented object detection is meaningful and gradually being emphasized.

---

*Corresponding author

Some current methods [11-13] model the oriented bounding box into Gaussian distributions (see Eq.1) and then use the distance measure in statistics (such as Wassertein Distance, Kullback-Leibler Divergence, and Bhattacharyya Distance) to calculate the similarity, effectively avoiding the boundary problems. However, there is no clear correspondence between the calculated similarity and SkewIoU. On this basis, Yang et al used Kalman filter to mimic the computing mechanism of SkewIoU, but the inconsistency between loss and evaluation metric still exists.

In this paper, we explore a more accurate modulated Kalman IoU of approximate SkewIoU for oriented object detection, named MKIoU. Through the analysis, we reveal that the inaccurate regression is due to the insensitivity of the loss variation tendency when the predicted box is close to the ground truth box, i.e., the inconsistency between the loss and the metric. Thus, fully considering the sensitivity of angle deviation and width-height offset to the loss variation, a modulation method is proposed to make the loss more consistent with the evaluation metric. Furthermore, the Gaussian modeling methods cannot learn the angle information of square targets, as shown in Fig 1 (top), which affects the performance of high-precision detection of the model. Thus, a GA Loss is proposed, which can correct the angle confusion of square objects, as shown in Fig 1 (down). Moreover, it can be easily added to other Gaussian modeling methods. The highlights of this paper can be summarized as follows:

1) A more accurate modulated Kalman IoU loss is proposed, named MKIoU, which is more concerned with the consistency of the loss trend with the evaluation metric. Compared to KFIoU[14], our method achieves higher accuracy.

2) A Gaussian Angle loss is presented to overcome the angle confusion problem of Gaussian modeling methods for square objects, thus further improving the performance of high-precision detection.

3) The proposed methods can be easily implemented in existing deep learning frameworks. Extensive experimental results on three public datasets show the effectiveness of our proposed methods, which achieve comparable performance to the state-of-the-art methods on the DOTA datasets.

## 2. Related Work

In this section, we briefly review existing works on oriented object detection.

### 2.1 Oriented Object Detector

Most of the existing oriented object detectors are based on mature horizontal object detectors such as Fast R-CNN[6], Faster R-CNN[15], YOLO[16,17], SSD[18], and RetinaNet[19]. Compared to general object detection, oriented object detection is mainly applied to aerial images and scene text. Arbitrary orientations, cluttered arrangements, and small objects are the main challenges for oriented object detection[2].

Some early methods [1,20,21] used multiple-angle anchors to detect oriented objects. However, the increase of anchors leads to a considerable increase in the calculation. RoI Transformer[3]proposed a rotated RoI (Region of Interest, RoI) learner, which can transform horizontal RoIs to rotated RoIs, avoiding a large number of rotated anchors. Gliding Vertex[4] proposed a new representation for oriented objects to avoid representation confusion. R$^3$Det[5] designs a feature refinement module by pixel-wise feature interpolation to obtain more accurate features. CSL[22] and DCL[23] regard angular prediction as a classification task. ReDet[24] proposes a Rotation-equivariant detector to extract rotation-equivariant features, which can accurately predict the direction. Oriented RepPoints[25]employs a set of adaptive points to capture the geometric, spatial, and semantic scenario. AOPG[26] first produces coarse-oriented boxes in an anchor-free manner and then refines them into high-quality-oriented proposals.

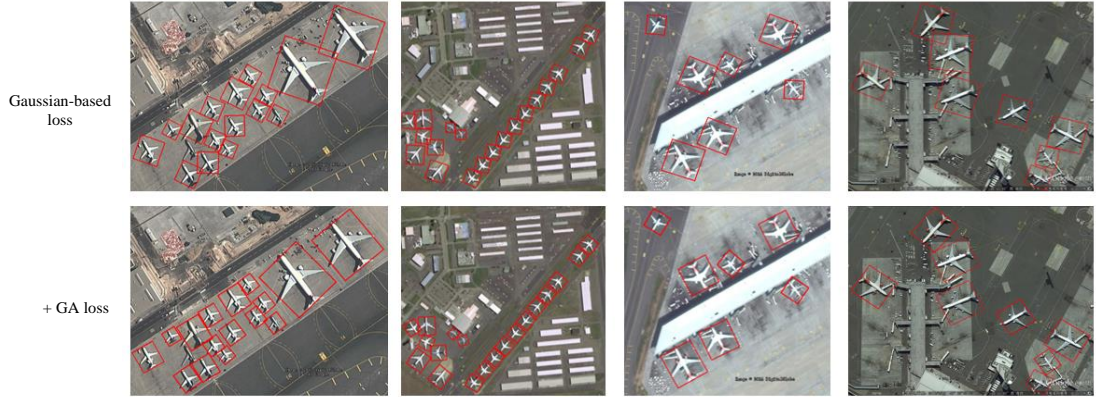

Fig 1. Visualization of detection results of Gaussian-based losses and adding Gaussian Angle Loss.

**2.2 Regression Losses**

The regression loss function is designed to continuously optimize the position of the predicted box to approach the ground truth box, thus directly determining the performance of the detector. For general object detection, regression loss has achieved relatively mature development, such as the commonly used Ln-norms losses (L1, L2, and SL1[6]) and IoU losses (GIoU[7], DIoU[8], CIoU[8], and EIoU[9]). However, in oriented object detection, the complex and variable nature of the oriented box makes the boundary problem prominent. Recently, some improved methods for loss function have been proposed.

**For Smooth L1 loss**, to avoid learning confusion caused by the sequence of the points, RSDet[27] proposes a modulated rotation loss, and SBD[28] proposes a sequential-free box discretization method. RIL[29] transforms the bounding box regression problem into an optimal matching problem, solving the boundary problem caused by ambiguous representations. SCRDet[2] adds the IoU constant factor to Smooth L1 loss to solve the boundary problem. On this basis, Area-IoU[30] introduces an area-guided function and IoU-guided function to solve the problems of scale imbalance and angle periodicity problems. **For approximate SkewIoU loss**, PIoU[31] approximates the intersection area using the number of pixels in the overlapping area. To tackle the uncertainty of convex caused by rotation, RIoU[32] defines a projection operation to estimate the intersection area. CFA[33] proposes a convex hull feature adaptation method based on CIoU loss for point-based detectors. In order to avoid the boundary problem of the oriented bounding box, GWD[11] first proposes to convert the oriented bounding box into a 2-D Gaussian distribution and then measures the loss by the Wasserstein distance between the two distributions, which effectively solves the boundary problem of the oriented box. However, when the object is close to a square, the angle information cannot be accurately learned. KLD[12], ProbIoU[13], and KFIoU[14] are also based on Gaussian modeling. However, the square problem still exists.

## 3. Method

In this section, we will describe the details of the proposed methods. We first analyze the inadequacy of the KFIoU Loss. Then a modulated Kalman IoU loss is proposed based on KFIoU, which is more concerned with the sensitivity of angular deviation and width-height offset to loss variation. Furthermore, Gaussian Angle loss is presented to overcome the angle confusion problem of square targets caused by Gaussian modeling methods.

### 3.1 Modulated Kalman Intersection over Union Loss

Firstly, to avoid the boundary problem, we follow the GWD[11] to convert the oriented bounding box $(x, y, w, h, \theta)$ into 2-D Gaussian distributions $N(\mu, \Sigma)$ as follows:

$$\Sigma = R \Lambda R^T$$
$$= \begin{pmatrix} \cos\theta & -\sin\theta \\ \sin\theta & \cos\theta \end{pmatrix} \begin{pmatrix} w^2/4 & 0 \\ 0 & h^2/4 \end{pmatrix} \begin{pmatrix} \cos\theta & \sin\theta \\ -\sin\theta & \cos\theta \end{pmatrix}$$
$$= \begin{pmatrix} \dfrac{w^2\cos^2\theta + h^2\sin^2\theta}{4} & \dfrac{w^2 - h^2}{4}\sin\theta\cos\theta \\ \dfrac{w^2 - h^2}{4}\sin\theta\cos\theta & \dfrac{w^2\sin^2\theta + h^2\cos^2\theta}{4} \end{pmatrix} \quad (1)$$
$$\mu = (x, y)^T$$

where R represents the rotation matrix, and $\Lambda$ represents the diagonal matrix of eigenvalues. Then we can easily calculate the area of the oriented bounding box by its covariance:

$$S = 4 \cdot |\Sigma|^{\frac{1}{2}} \quad (2)$$

But the key is how to calculate the intersection area. Inspired by KFIoU[14], we use Smooth L1 loss to make the center points of two Gaussian distributions coincide, then use the Kalman filter to obtain the covariance $\Sigma_{pt}$ of overlapping regions of two Gaussian distributions:

$$\Sigma_{pt} = \Sigma_p - \Sigma_p(\Sigma_p + \Sigma_t)^{-1}\Sigma_p \quad (3)$$

where $\Sigma_p$ is the covariance matrix of the predicted box, $\Sigma_t$ is the covariance matrix of the ground truth box. Combined with Eq. 2, we can obtain the approximate area $S_{pt}$ of the intersection of the two oriented bounding boxes. Then, the KFIoU can be expressed as:

$$\text{KFIoU} = \frac{S_{pt}}{S_p + S_t - S_{pt}} \quad (4)$$

The KFIoU Loss is easy to implement in current deep learning frameworks. However, the result of KFIoU loss in the higher precision index ($AP_{75}$ and $AP_{50:95}$) is not good, i.e., the regression results are not accurate. By combining Eq. 1 to Eq. 4, we have:

$$\text{KFIoU} = \frac{1}{A + B - 1} \quad (5)$$

$$A = \sqrt{1 + \frac{w_p^2 h_p^2}{w_t^2 h_t^2} + (\frac{w_p^2}{w_t^2} + \frac{h_p^2}{h_t^2})\cos^2\!\Delta\theta + (\frac{w_p^2}{h_t^2} + \frac{h_p^2}{w_t^2})\sin^2\!\Delta\theta} \quad (6)$$

$$B = \sqrt{1 + \frac{w_t^2 h_t^2}{w_p^2 h_p^2} + (\frac{w_t^2}{w_p^2} + \frac{h_t^2}{h_p^2})\cos^2\!\Delta\theta + (\frac{w_t^2}{h_p^2} + \frac{h_t^2}{w_p^2})\sin^2\!\Delta\theta} \quad (7)$$

It can be obtained that $A + B \geq 4$, the equation holds when one of the following conditions is satisfied:

$$\begin{cases} \theta = 0°, & w_p = w_t, h_p = h_t \\ \theta = 90°, & w_p = h_t, w_t = h_p \end{cases} \tag{8}$$

That is, when the predicted box overlaps completely with the ground-truth box, the KFIoU takes the maximum values of 1/3. We do a visualization analysis of KFIoU, as shown in Fig 2. Note that for convenience comparison, we multiply KFIoU by 3 so that the range of values is [0,1]. It can be found that there is a significant difference between the variation trend of KFIoU (red) and the actual trend of SkewIoU(blue). Specifically, when the predicted box is close to the ground truth box, KFIoU is much larger than the actual IoU, which makes the optimization of the model insensitive, thus damaging the high-precision detection performance of the model. We expect that the variation trend of the approximate IoU to be consistent with SkewIoU. Therefore, we propose MKIoU, which is adjusted by a modulation factor α:

$$\mathrm{MKIoU} = \frac{4-\alpha}{A+B-\alpha}, \ \alpha < 4 \tag{9}$$

The value range of MKIoU is (0,1]. KFIoU is a special case of MKIoU, when α=1, MKIoU is equal to $3 \cdot$ KFIoU. Fig 2(a) shows the sensitivity of the width-height offset of MKIoU for different values of α when the two bounding boxes have the same center and angle. It can be seen that when α=1, the sensitivity of MKIoU to width-height offset is much lower than that of SkewIoU, which leads to inaccurate results. As α increases, the sensitivity increases gradually, making the trend of MKIoU more consistent with the evaluation metric. Fig 2(b) shows the sensitivity of the angle deviation of MKIoU when the two bounding boxes have the same center, width, and height. Similar conclusions can be drawn to the above. Note that with the increase of the aspect ratio of the object, the sensitivity of the SkewIoU to the angle increases, and a small angle change will also cause a sharp change in SkewIoU, which is not conducive to a stable and accurate regression for the loss function. Therefore, α is not the bigger the better, the final value of α is determined by comparative experiments.

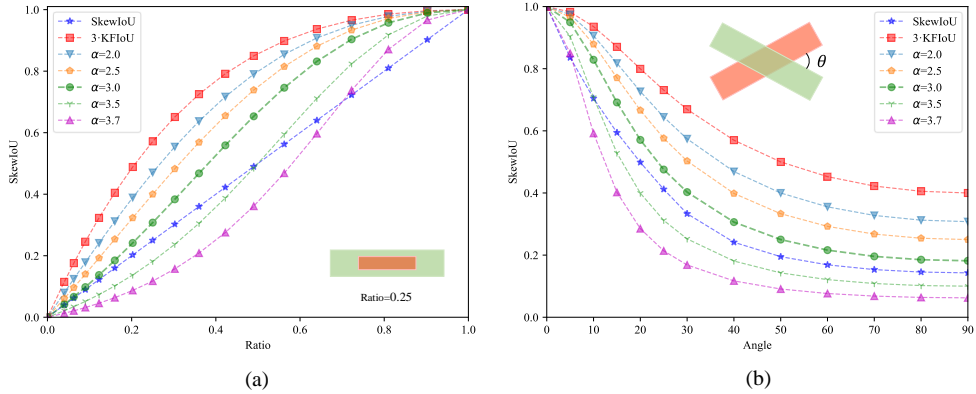

Fig 2. Comparison of MKIoU curves under different conditions

### 3.2 Gaussian Angle Loss for angle confusion problem of Gaussian modeling

The Gaussian-based approaches effectively avoid the boundary problem, but when the object is a square, i.e., $w=h$, its 2D Gaussian distribution is a circle that does not accurately represent the orientation. Specifically, when $w=h$, then

$$\Sigma = \begin{pmatrix} \frac{w^2}{4} & 0 \\ 0 & \frac{w^2}{4} \end{pmatrix} \tag{10}$$

The angle information is missing. This will lead to the angle confusion of square-like objects, which will affect the performance of high-precision detection. Fig. 1 (top) shows the visualization of angle confusion problem. To solve this problem, we propose the Gaussian Angle loss, which can be expressed as:

$$L_{GA} = \beta e^{4\lambda - \lambda(\frac{w_t}{h_t} + \frac{h_t}{w_t})^2} \sin^2(2\Delta\theta) \tag{11}$$

where $\Delta\theta = \theta_p - \theta_t$, $\lambda$ and $\beta$ are hyperparameters. To avoid impact on non-square objects, we set $\lambda = 3$ by default. The correction of the $\theta_P$ is only related to the angle deviation and the aspect ratio of the ground-truth. For further analysis, we have:

$$\frac{\partial L_{GA}}{\partial \theta_P} = 2 e^{4\lambda - \lambda(\frac{w_t}{h_t} + \frac{h_t}{w_t})^2} \sin(4\Delta\theta) \tag{12}$$

when $w_t = h_t$, we have $\partial L_{GA}/\partial \theta_P = 2\sin(4\Delta\theta)$. This means that as the aspect ratio of ground-truth approaches 1, the angle gradient will show a periodic change $90°$, which is consistent with the angular period of the square. As the aspect ratio increases, $\partial L_{GA}/\partial \theta_P$ is closed to 0, and the GA loss will be invalidated. The 3D diagram of $L_{GA}$ is shown in Fig 3. It more intuitively expresses the above characteristics. GA Loss can be readily implemented on MKIoU and easily added to other Gaussian modeling methods, and the performance of high-precision detection is improved.

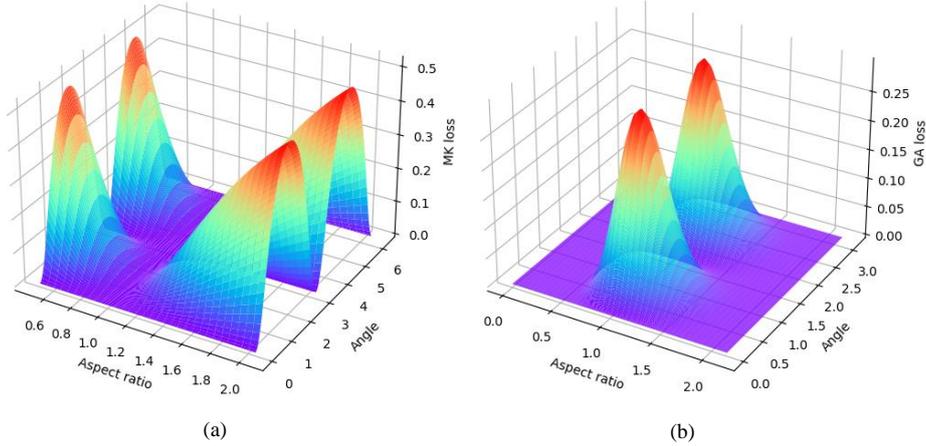

Fig 3 The relationship between loss (Z-axis), aspect ratio (X-axis), and angle deviation (Y-axis). The MK Loss (a) varies $180°$ periodically with the change of angle deviation. The larger the aspect ratio, the greater the loss, indicating the model pays more attention to optimizing objects with a larger aspect ratio. However, as the aspect ratio approaches 1, the MK Loss approaches 0, and there is no gradient change. The GA Loss (b) varies by $90°$ periodically with the change of angle deviation when the aspect ratio approaches 1. As the aspect ratio becomes larger, the loss approaches 0, thus not affecting the original loss value.

### 3.3 Advanced detectors with the proposed loss

Our final regression loss can be expressed as:

$$L_{reg} = \sum_{i \in (x,y)} L_s(t_i, t_i^*) + L_{MKIoU} + L_{GA} \tag{13}$$

where $L_s$ is Smooth L1 Loss, used for the regression of the center point. $L_{MKIoU} = 1 - \text{MKIoU}$. $L_{GA}$ is Gaussian Angle loss, used to overcome the angle confusion problem of Gaussian modeling methods for square objects.

Our methods can be easily applied to the current advanced detectors on MMrotate[37], such as Rotated RetinaNet, R³Det[5], and RoI Transformer[3]. In this paper, we configure our loss on all three detectors and do corresponding experiments. The multi-task loss is:

$$L_{total} = \lambda_1 \sum_{n=1}^{N_{pos}} L_{reg}(p_n, t_n) + \frac{\lambda_2}{N} \sum_{n=1}^{N} L_{cls}(lp_n, lt_n) \quad (14)$$

where $\lambda_1$ and $\lambda_2$ are hyperparameters and set to {0.01,1}. $N_{pos}$ and $N$ represent the number of when positive anchors and all anchors, respectively. $p_n$ indicate the $n$-th predicted bounding box, $t_n$ is the $n$-th ground-truth box. $lp_n$ is the $n$-th probability distribution of various classes calculated by a sigmoid function. $lt_n$ represents the label of the $n$-th object. For Rotated RetinaNet and R³Det, we use Focal loss[19] as classification loss $L_{cls}$, and for RoI Transformer, we use Cross Entropy Loss.

## 4. Experiments

In this section, experiments are constructed to evaluate the performance of our MKIoU loss.

### 4.1 Datasets

Our experiments evaluate on DOTA1.0[34], UCAS_AOD[35], and HRSC2016[36] datasets.

**DOTA** is currently the most commonly used dataset for oriented object detection and one of the most challenging datasets in aerial images. DOTA1.0 contains 2,806 images and 188,282 instances of 15 categories: plane (PL), baseball diamond (BD), bridge (BR), ground field track (GFT), small vehicle (SV), large vehicle (LV), ship (SH), tennis court (TC), baseball curt (BC), storage tank (ST), soccer-ball field (SBF), roundabout (RA), harbor (HA), swimming pool (SP) and helicopter (HC). The image size ranges from around 800×800 to 4,000×4,000. We divide the images into 1024×1024 sub-images with an overlap of 200, and the multi-scale argument of DOTA1.0 are {0.5,1.0,1.5}

**UCAS_AOD** is a car and airplane dataset that contains 1,510 images with 14,596 instances. We randomly select 1,057 images for training and 453 images for testing.

**HRSC2016** is a ship dataset that contains 1061 images with 2976 instances. The image size ranges from 300 × 300 to 1,500 × 900 pixels. We randomly select 617 images for training and 444 images for testing. All images were resized to 800 × 800 for training and testing.

### 4.2 Implement details

We use MMrotate[37] to conduct all our experiments where many advanced rotation detectors are integrated. We use Pytorch[38] with 1 GeForce RTX 2080 Ti for single-scale training and 1 NVIDIA A100 for multi-scale training. Experiments are initialized by ResNet50[39] by default unless otherwise specified. We use SGD optimizer to train the model. The momentum and weight decay are 0.9 and 0.0001, respectively. The batchsize is 2 (2 images per GPU). The initial learning rate is 2.5e-3 by default. For DOTA datasets, we trained 12 epochs in total, and the learning rate is reduced by a factor of 10 at 8 epochs and 11 epochs, respectively, and doubled if data augmentation or multi-scale training is used. For UCAS_AOD, we trained 12 epochs in total, and the learning rate is reduced by a factor of 10 at 8 epochs and 11 epochs, respectively. For HRSC2016, we trained 72 epochs in total, and the learning rate is 1.25e-3 and reduced by a factor of 10 at 48 epochs and 66 epochs.

### 4.3 Evaluation metrics

mAP is the most critical metric for evaluating the performance in object detection. It can be divided into COCO[40] metric and VOC[41] metric. The mAP of the COCO metric uses 10 IoU thresholds 0.5:0.05:0.95. First, the AP of all classes under each threshold are calculated, then mAP is obtained by averaging the AP under 10 thresholds. While the VOC metric only considers one IoU threshold, i.e., IoU = 0.5. In this paper, unless otherwise specified, we use the COCO metric to compare the performance of high precision detection more accurately.

### 4.4 Ablation study

**1)** hyperparameter experiments

We perform detailed hyperparameter experiments on the proposed loss in this paper. Table 1 provides the results of the experiments under different α and β on RetinaNet. Note that KF Loss means 1-KFIoU, $f(KFIoU)$ refers to the nonlinear transformation. Follow the paper[14], $f(KFIoU) = e^{(1-KFIoU)} - 1$ for RetinaNet, and $f(KFIoU) = -\ln(KFIoU)$ for R$^3$Det and RoI Transformer. Due to the label of the Test set of DOTA is not public, this set of experiments was trained on the training set and tested on the validation set for convenience. The performance reaches the optimal when α=3, β=0.3. Unless otherwise specified, we will keep this parameter configuration in the following experiments. * indicates that the results were implemented on MMrotate.

Table 1. Ablation study of different hyperparameters on DOTA_v1.0

| Method | Loss | | Train/val | | | |
|---|---|---|---|---|---|---|
| | | | $AP_{50}$ | $AP_{75}$ | $AP_{85}$ | mAP |
| RetinaNet | KFLoss* | | 64.53 | 28.19 | 10.02 | 33.00 |
| | $f(KFIoU)$* | | 64.63 | 29.27 | 12.14 | 34.34 |
| | MKIoU | α=2.6 | 64.19 | 33.53 | 15.13 | 35.64 |
| | | α=2.8 | 64.22 | 33.12 | 14.43 | 35.39 |
| | | **α=3.0** | **63.94** | **33.83** | **14.70** | **35.76** |
| | | α=3.2 | 63.91 | 32.77 | 15.40 | 35.59 |
| | | α=3.4 | 62.79 | 32.91 | 15.74 | 35.40 |
| | MKIoU +GA | β=0.1 | 63.20 | 32.73 | 14.24 | 35.23 |
| | | β=0.2 | 64.23 | 34.25 | 15.29 | 35.95 |
| | | **β=0.3** | **64.39** | **34.93** | **16.12** | **36.66** |
| | | β=0.4 | 63.42 | 34.73 | 15.27 | 35.93 |
| | | β=0.5 | 63.16 | 34.70 | 15.45 | 35.53 |

**2)** Ablation study of the proposed methods on different datasets

Table 2 shows the ablation experiments of the proposed method on UCAS_AOD. We performed experiments on two oriented detectors, RetinaNet and R$^3$Det. It can be seen from the table that the proposed MKIoU is better than KFLoss and $f(KFIoU)$ in terms of high precision, which indicates that adjusting width-height sensitivity and angle sensitivity is beneficial to more accurate regression. Furthermore, the proposed GA loss further improved the accuracy. Combined with Fig 2, we can conclude that the GA loss can effectively correct the angle confusion problem of square objects caused by Gaussian modeling. It should be noted that the number of planes accounts for half in UCAS_AOD and its bounding box is close to square. Therefore, the high-precision results of Gaussian-based methods on this dataset are poor.

Table 2. Ablation study on UCAS_AOD

| Method | Loss | $AP_{50}$ | $AP_{75}$ | mAP |
|---|---|---|---|---|
| RetinaNet | KFLoss* | 89.90 | 29.50 | 45.78 |
| | $f(KFIoU)$ * | 89.89 | 29.01 | 44.69 |
| | **MKIoU** | **89.80** | **36.49** | **47.73** |
| | **MKIoU+GA** | **89.88** | **55.59** | **53.85** |
| R$^3$Det | KFLoss* | 89.89 | 21.82 | 40.42 |
| | $f(KFIoU)$ * | 90.20 | 34.29 | 47.82 |
| | **MKIoU** | **90.27** | **40.22** | **48.39** |
| | **MKIoU+GA** | **90.13** | **50.33** | **52.23** |

Table 3 Ablation study on HRSC2016

| Method | Loss | $AP_{50}$ | $AP_{75}$ | $AP_{85}$ | mAP |
|---|---|---|---|---|---|
| RetinaNet | Smooth L1 | 84.28 | 48.42 | 12.56 | 47.76 |
| | GWD | 85.56 | 60.31 | 17.14 | 52.89 |
| | KLD | 87.45 | 72.39 | 27.68 | 57.80 |
| | KFIoU | 84.41 | 58.32 | 18.34 | 51.29 |
| | **MKLoss(ours)** | **85.98** | **70.16** | **26.23** | **57.02** |
| $R^3$Det | Smooth L1 | 88.52 | 43.42 | 4.58 | 46.18 |
| | GWD | 89.43 | 65.88 | 15.02 | 56.07 |
| | KLD | 89.97 | 77.38 | 25.12 | 61.40 |
| | KFIoU* | 90.20 | 64.78 | 20.76 | 56.70 |
| | **MKLoss(ours)** | **90.13** | **76.65** | **29.80** | **61.12** |

Table 3 shows the high-precision experimental comparison of our method with other Gaussian-based losses on HRSC2016. GWD and KLD are currently advanced loss functions in high-precision detection, which is also based on Gaussian modeling. From the table, we can see that KFIoU can achieve comparable results with GWD, but there is still a big gap with KLD. We analyze that this is due to the targets in HRSC2016 are ships with a large aspect ratio, while KFIoU is insensitive to angle changes. To make the loss more sensitive to angle changes, we increase the α, but being too sensitive will lead to unstable model training, so we finally set α=2.5 in HRSC2016. Our method greatly improves the detection accuracy (On RetinaNet, 11.84%, 7.89%, 5.73% improvement on $AP_{75}$, $AP_{85}$, and mAP, respectively) and achieves comparable results to KLD. Similar results can be concluded in $R^3$Det.

To better demonstrate the effectiveness of the proposed methods, we conducted ablation experiments on the DOTAv1.0 dataset with two different detectors, as shown in Table 4. The result of $f$ (KFIoU) is better than KFLoss, but the high precision index is still not well. Our proposed MKIoU maintains comparable performance on $AP_{50}$ and shows a significant improvement on $AP_{75}$ and mAP, with an increase of 1.34% and 0.73% on RetinaNet, respectively. After adding GA Loss, $AP_{75}$ and mAP are further improved by 2.37% and 1.00%, respectively. Even on the better detectors (RoI Transformer), our method still improves the high precision detection performance of the model.

Table 4. Ablation study on DOTA

| Method | Loss | Trainval/test | | |
|---|---|---|---|---|
| | | $AP_{50}$ | $AP_{75}$ | mAP |
| RetinaNet | KFLoss* | 69.26 | 31.66 | 36.31 |
| | $f$ (KFIoU)* | 69.60 | 36.64 | 38.23 |
| | **MKIoU** | **68.99** | **37.98** | **38.90** |
| | **MKIoU+GA** | **69.40** | **40.35** | **39.90** |
| RoI Transformer | KFIoU* | 75.16 | 39.33 | 41.12 |
| | $f$ (KFIoU)* | 75.35 | 40.73 | 41.94 |
| | **MKIoU** | **76.16** | **42.63** | **42.75** |
| | **MKIoU+GA** | **75.90** | **43.85** | **43.74** |

Table 5. Ablation study for GA Loss under Gaussian-based losses on UCAS_AOD

| Method | Loss | $AP_{50}$ | $AP_{75}$ | mAP |
|---|---|---|---|---|
| RetinaNet | Smooth L1* | 89.74 | 46.22 | 49.40 |
| | GWD* | 89.80 | 36.10 | 47.61 |
| | **GWD+GA** | **89.94** | **56.16** | **54.23** |
| | KLD* | 89.85 | 36.29 | 48.04 |
| | **KLD+GA** | **89.92** | **54.52** | **53.51** |
| $R^3$Det | Smooth L1* | 89.69 | 42.69 | 47.55 |
| | GWD* | 90.10 | 36.01 | 48.09 |
| | **GWD+GA** | **90.16** | **56.63** | **54.05** |
| | KLD* | 90.23 | 42.15 | 49.48 |
| | **KLD+GA** | **90.09** | **53.01** | **52.85** |

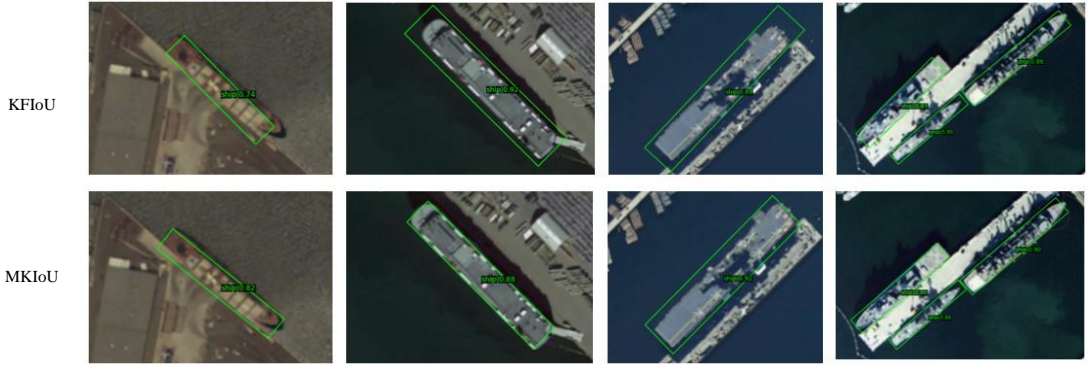

Fig 4. Visual comparison between KFIoU(top) and MKIoU(down) on HRSC2016.

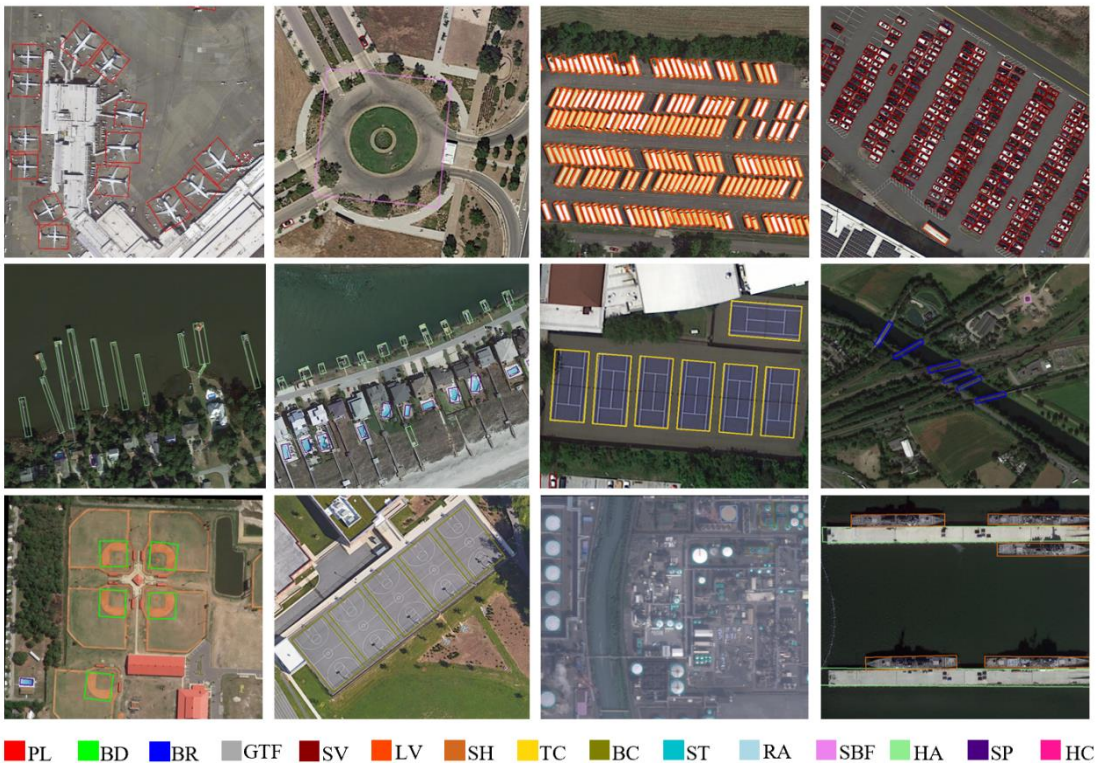

Fig 5. Visualization of detection results on the DOTA dataset

**The generality of Gaussian Angle Loss.** Table 5 shows the comparison results of other Gaussian-based losses (KLD and GWD) after adding GA loss under different detectors on UCAS_AOD. It can be seen from the table that the proposed Gaussian angle loss is equally effective for KLD and GWD and dramatically improves the detection accuracy, which suggests that the GA Loss is generalizable for Gaussian-based loss.

3) Comparison with the State-of-the-Arts Methods

**Results on HRSC2016.** Table 6 shows the comparison of the proposed method with the other advanced methods on HRSC2016 datasets. The proposed MKIoU achieves the best results 90.45% (VOC 2007 metric) on ResNet 101, which indicates that our method is effective.

**Results on DOTA.** To further demonstrate the robustness of the proposed method, we extensively compare the proposed method with other advanced methods on the more challenging DOTA dataset. As shown in Table 7, our method achieves 77.76% mAP in single-scale strategy and 80.83% mAP in multi-scale strategy, which is comparable to the current state-of-the-arts methods. Compared with KFIoU, the $AP_{50}$ of our method is 0.6% higher. The visualization of detection results is shown in Fig 5. Furthermore, Table 8 shows the high-precision results on DOTAv1.0. In single-scale strategy, our method is slightly higher than KFIoU on $AP_{50}$ but significantly better on $AP_{75}$ and $AP_{50:95}$, with an increase of 1.71% and 0.78%, respectively. This fully shows that our method has more advantages in high-precision detection. In multi-scale strategy, our method achieves 80.83%, 55.46%, and 51.46% mAP on $AP_{50}$, $AP_{75}$, and $AP_{50:95}$, respectively.

Table 6. Comparison of the mAP (VOC 07 metric) of different methods on HRSC2016

| Method | Backbone | mAP (%) |
|---|---|---|
| TOSO[42] | ResNet-101 | 79.29 |
| RoI-Trans[3] | ResNet-101 | 86.20 |
| Gliding Vertex[4] | ResNet-101 | 88.20 |
| BBAVectors[43] | ResNet-101 | 88.60 |
| $R^3$Det[5] | ResNet-101 | 89.26 |
| $R^3$det-DCL[23] | ResNet-101 | 89.46 |
| FPN-CSL[22] | ResNet-101 | 89.62 |
| DAL[44] | ResNet-101 | 89.77 |
| $R^3$Det-GWD[11] | ResNet-101 | 89.85 |
| $S^2$ANet[45] | ResNet-101 | 90.17 |
| $S^2$ANet-SASM[46] | ResNet-101 | 90.27 |
| **Ours** | ResNet-101 | **90.45** |

Table 7. Comparison with advanced methods on the DOTA dataset. Here R-50, R-101, R152 represent ResNet-50, ResNet101, ResNet152, respectively. RX-101, ReR-50 denote ResNeX101[47] and ReResNet-50, respectively. MS indicates that multi-scale training and testing is used.

| Methods | Backbone | MS | PL | BD | BR | GTF | SV | LV | SH | TC | BC | ST | SBF | RA | HA | SP | HC | $AP_{50}$ |
|---|---|---|---|---|---|---|---|---|---|---|---|---|---|---|---|---|---|---|
| RoI-Trans[3] | R-101 | √ | 88.64 | 78.52 | 43.44 | 75.92 | 68.81 | 73.68 | 83.59 | 90.74 | 77.27 | 81.46 | 58.39 | 53.54 | 62.83 | 58.93 | 47.67 | 69.56 |
| P-RSDet[48] | R-101 | √ | 88.58 | 77.83 | 50.44 | 69.29 | 71.10 | 75.79 | 78.66 | 90.88 | 80.10 | 81.71 | 57.92 | 63.03 | 66.30 | 69.77 | 63.13 | 72.30 |
| SCRDet[2] | R-101 | √ | 89.98 | 80.65 | 52.09 | 68.36 | 68.36 | 60.32 | 72.41 | 90.85 | 87.94 | 86.86 | 65.02 | 66.68 | 66.25 | 68.24 | 65.21 | 72.61 |
| CFC-Net[49] | R-101 | √ | 89.08 | 80.41 | 52.41 | 70.02 | 76.28 | 78.11 | 87.21 | 90.89 | 84.47 | 85.64 | 60.51 | 61.52 | 67.82 | 68.02 | 50.09 | 73.50 |
| Li et.al[30] | R-101 | √ | 88.69 | 79.41 | 52.26 | 65.51 | 74.72 | 80.83 | 87.42 | 90.77 | 84.31 | 83.36 | 62.64 | 58.14 | 66.95 | 72.32 | 69.34 | 74.44 |
| GV[4] | R-101 | | 89.64 | 85.00 | 52.26 | 77.34 | 73.01 | 73.14 | 86.82 | 90.74 | 79.02 | 86.81 | 59.55 | 70.91 | 72.94 | 70.86 | 57.32 | 75.02 |
| MASK[50] | RX-101 | √ | 89.56 | 85.95 | 54.21 | 76.52 | 74.16 | 75.63 | 85.63 | 89.85 | 83.81 | 86.48 | 54.89 | 69.64 | 73.94 | 69.06 | 63.32 | 75.33 |
| $R^3$Det[5] | R-152 | √ | 89.80 | 83.77 | 48.11 | 66.77 | 78.76 | 83.27 | 87.84 | 90.82 | 85.38 | 85.51 | 65.67 | 62.68 | 67.53 | 78.56 | 72.67 | 76.47 |
| DAL[44] | R-50 | √ | 89.69 | 83.11 | 55.03 | 71.00 | 78.30 | 81.90 | 88.46 | 90.89 | 84.97 | 87.46 | 64.41 | 65.56 | 76.86 | 72.09 | 64.35 | 76.95 |
| SCRDet++[51] | R-101 | √ | 90.05 | 84.39 | 55.44 | 73.99 | 77.54 | 71.11 | 86.05 | 90.67 | 87.32 | 87.08 | 69.62 | 68.90 | 73.74 | 71.29 | 65.08 | 76.81 |
| DCL[23] | R-152 | √ | 89.26 | 83.60 | 53.54 | 72.76 | 79.04 | 82.56 | 87.31 | 90.67 | 86.59 | 86.98 | 67.49 | 66.88 | 73.29 | 70.56 | 69.99 | 77.37 |
| AProNet[52] | R-101 | √ | 88.77 | 84.95 | 55.27 | 78.40 | 76.65 | 78.54 | 88.45 | 90.83 | 86.56 | 87.01 | 65.62 | 70.29 | 75.43 | 78.17 | 67.28 | 78.16 |
| $S^2$ANet[53] | R101 | √ | 89.28 | 84.11 | 56.95 | 79.21 | 80.18 | 82.93 | 89.21 | 90.86 | 84.66 | 87.61 | 71.66 | 68.23 | 78.58 | 78.20 | 65.55 | 79.15 |
| $O^2$DETR[54] | R-50 | √ | 88.89 | 83.41 | 56.72 | 79.75 | 79.89 | 85.45 | 89.77 | 90.84 | 86.15 | 87.66 | 69.94 | 68.97 | 78.83 | 78.19 | 70.38 | 79.66 |
| SARA[46] | R-50 | √ | 89.40 | 84.29 | 56.72 | 82.29 | 80.49 | 83.01 | 89.37 | 90.67 | 86.20 | 87.44 | 71.34 | 69.06 | 78.49 | 80.98 | 68.98 | 79.91 |
| ReDet[24] | ReR-50 | √ | 88.81 | 82.48 | 60.83 | 80.82 | 78.34 | 86.06 | 88.31 | 90.87 | 88.77 | 87.03 | 68.65 | 66.90 | 79.26 | 79.71 | 74.67 | 80.10 |
| GWD[11] | R-152 | √ | 89.66 | 84.99 | 59.26 | 82.19 | 78.97 | 84.83 | 87.70 | 90.21 | 86.54 | 86.85 | 73.47 | 67.77 | 76.92 | 79.22 | 74.92 | 80.23 |
| KLD[12] | R-152 | √ | 89.92 | 85.13 | 59.19 | 81.33 | 78.82 | 84.38 | 87.50 | 89.80 | 87.33 | 87.00 | 72.57 | 71.35 | 77.12 | 79.34 | 78.68 | 80.63 |
| KFIoU[14] | R-50 | √ | 89.12 | 84.54 | 60.73 | 78.86 | 79.65 | 85.79 | 88.45 | 90.90 | 87.03 | 88.28 | 69.15 | 70.28 | 78.88 | 81.54 | 70.05 | 80.23 |
| OURS | R-50 | | 89.36 | 83.93 | 57.74 | 71.93 | 79.25 | 84.86 | 88.18 | 90.88 | 88.27 | 86.24 | 64.89 | 63.95 | 77.99 | 73.73 | 65.24 | 77.76 |
| OURS | R-50 | √ | 89.30 | 85.14 | 60.64 | 79.99 | 80.40 | 85.88 | 88.47 | 90.89 | 87.26 | 87.86 | 70.10 | 72.63 | 79.26 | 81.66 | 72.91 | 80.83 |

Table 8. High-precision results on DOTAv1.0

| Methods | Backbone | MS | $AP_{50}$ | $AP_{75}$ | $AP_{50:95}$ |
|---|---|---|---|---|---|
| KFIoU* | R-50 | | 77.73 | 47.00 | 46.34 |
| OURS | R-50 | | 77.76 | 48.71 | 47.12 |
| OURS | R-50 | √ | 80.83 | 55.46 | 51.46 |

# 5. Conclusion

In this paper, we proposed a modulated Kalman IoU Loss of approximate SkewIoU for oriented object detection, named MKIoU Loss. Compared with KFIoU, MKIoU pays more attention to the sensitivity of width-height offset and angle deviation to loss variation, thus significantly improving the high-precision detection performance of the model. Furthermore, Gaussian Angle loss is presented to overcome the angle confusion problem for square objects of the Gaussian modeling method. Experimental on several public datasets demonstrate the effectiveness of our approach and achieve comparable results to state-of-the-art methods. Although our method achieves more accurate detection, there are still shortcomings, such as misclassification and detection of objects with huge aspect ratios. Therefore, in future work, we will focus on effectively reducing misclassification and improving objects' precision with huge aspect ratios.